%% file: main.tex
\documentclass[11pt,a4paper]{article}
\usepackage[usenames, dvipsnames]{color}
\usepackage[nohyperref]{acl18-latex/acl2018}
\usepackage{times}
\usepackage{latexsym}
\usepackage{multirow}
\usepackage{array}
\usepackage{amsmath}
\usepackage{graphicx}
\usepackage{quoting}

\quotingsetup{vskip=3pt}

\usepackage{url}

\aclfinalcopy 
\def\aclpaperid{20} 

\newcommand{\dataset}[0]{BTRIS-Mobility}

\newcommand{\BTRIS}[0]{BTRIS}
\newcommand{\PTOT}[0]{PT-OT}
\newcommand{\Mob}[0]{Mobility}
\newcommand{\ScoreDef}[0]{ScoreDefinition}

\newcommand{\MobColor}[1]{\textcolor{RedOrange}{#1}}
\newcommand{\SDColor}[1]{\textcolor{Cerulean}{#1}}

\newcolumntype{P}[1]{>{\centering\arraybackslash}p{#1}} 
\newcolumntype{M}[1]{>{\centering\arraybackslash}m{#1}} 

\title{Embedding Transfer for Low-Resource Medical Named Entity Recognition: A Case Study on Patient Mobility}

\author{Denis Newman-Griffis$^{1,2}$ \and Ayah Zirikly$^1$ \\
  $^1$Rehabilitation Medicine Department, Clinical Center, National Institutes of Health, Bethesda, MD\\
  $^2$Department of Computer Science and Engineering, The Ohio State University, Columbus, OH\\
  {\tt \{denis.griffis, ayah.zirikly\} @nih.gov }
}

\date{}

\begin{document}
\maketitle
\begin{abstract}
    \input{abstract}
\end{abstract}

\input{sections/introduction}
\input{sections/related-work}
\input{sections/data}
\input{sections/methods}
\input{sections/results}
\input{sections/qualitative}
\input{sections/conclusions}
\input{sections/acknowledgments}

\bibliographystyle{acl18-latex/acl_natbib}
\bibliography{references}

\end{document}

%% file: abstract.tex
Functioning is gaining recognition as an important indicator of global health,
but remains under-studied in medical natural language processing research. We present
the first analysis of automatically extracting descriptions of
patient mobility, using a recently-developed dataset of free text electronic
health records. We frame the task as a named entity recognition (NER) problem,
and investigate the applicability of NER techniques to mobility extraction.
As text corpora focused on patient functioning
are scarce, we explore domain adaptation of word embeddings for use in a
recurrent neural network NER system. We find that embeddings trained on a
small in-domain corpus perform nearly as well as those learned from large
out-of-domain corpora, and that domain adaptation techniques yield additional
improvements in both precision and recall. Our analysis
identifies several significant challenges in extracting descriptions of patient
mobility, including the length and complexity of annotated entities and
high linguistic variability in mobility descriptions.

%% file: sections/introduction.tex
\section{Introduction}

Functioning has recently been recognized as a leading world health indicator,
joining morbidity and mortality \cite{Stucki2017b}. Functioning is defined in
the
International Classification of Functioning, Disability, and Health (ICF; \citealt{ICF})
as the interaction between health conditions, body functions and structures,
activities and participation, and contextual factors.
Understanding functioning is an important element in assessing quality of life,
and automatic extraction of patient functioning would serve as a useful tool
for a variety of care decisions, including rehabilitation and disability
assessment \cite{Stucki2017c}.
In healthcare data,
natural language processing (NLP) techniques have been successfully used for
retrieving information about health conditions, symptoms
and procedures from unstructured electronic health record (EHR) text
\cite{Soysal2018,Savova2010}. As recognition
of the importance of functioning grows, there is a need to investigate
the application of NLP methods to other elements of functioning.

Recently, \newcite{Thieu2017} introduced a dataset of EHR documents
annotated for descriptions of patient mobility status, one area of activity
in the ICF. Automatically recognizing these descriptions faces significant
challenges, including their length and syntactic complexity and a lack of
terminological resources to draw on.  In this study, we view this task
through the lens of named entity recognition (NER), as
recent work has illustrated
the potential of using recurrent neural network (RNN) NER models to address
similar issues in biomedical NLP \cite{Xia2017,Dernoncourt2017a,Habibi2017}.

An additional strength of RNN models is their ability to
leverage pretrained word embeddings, which capture co-occurrence
information about words from large text corpora. Prior work has shown
that the best improvements come from embeddings
trained on a corpus related to the target domain
\cite{Pakhomov2016}. However, free text describing patient functioning
is hard to come by: for example, even the large MIMIC-III 
corpus \cite{Johnson2016} includes only a few hundred documents from therapy
disciplines among its two million notes. While recent work suggests that
using a training corpus from the target domain can mitigate a lack
of data \cite{Diaz2016}, even a careful corpus selection may
not produce sufficient data to train robust word representations.

In this paper, we explore the use of an RNN model to
recognize descriptions of patient mobility. We analyze the impact of
initializing the model with word embeddings trained on a variety of corpora,
ranging from large-scale
out-of-domain data to small, highly-targeted in-domain documents. We further
explore several domain adaptation techniques for combining word-level
information from both of these data sources, including a novel
nonlinear embedding transformation method using a deep neural network.

We find that embeddings trained on a very small set of therapy encounter
notes nearly match the mobility NER performance of representations trained on 
millions of out-of-domain documents. Domain adaptation of input word embeddings
often improves performance
on this challenging dataset, in both precision and recall. Finally, we find
that simpler adaptation methods such as concatenation and preinitialization
achieve highest overall performance, but that nonlinear mapping of embeddings
yields the most consistent performance across experiments. We achieve a
best performance of~69\% exact match and over 83\% token-level match F-1 score
on the mobility data, and identify
several trends in system errors that suggest fruitful directions for further
research on recognizing descriptions of patient functioning.

%% file: sections/related-work.tex
\section{Related work}
The extraction of named entities in free text has been one of the most
important tasks in NLP and information extraction (IE). As a result,
this track of research has matured over the last two decades, especially
in the newswire domain for high resource languages such as English. Many
of the successful existing NER systems use a combination of engineered
features trained using conditional random fields (CRF)
model~\cite{mccallum2003early,finkel2005incorporating}. NER systems have
also been widely studied in medical NLP, using dictionary lookup
methods~\cite{Savova2010}, support vector
machine (SVM) classifiers~\cite{kazama2002tuning}, and sequential
models \cite{tsai2006nerbio, settles2004biomedical}. 
In recent years, deep learning models have been used in NER with successful
results in many domains~\cite{collobert2011natural}. Proposed neural network
architectures included hybrid convolutional neural network (CNN) and
bi-directional long-short term memory (Bi-LSTM) as introduced by~\newcite{chiu2015named}.
State-of-the-art NER models use the architecture proposed by \newcite{lample2016neural}, 
a stacked bi-directional long-short term memory (Bi-LSTM) for both character and word,
with a CRF layer on the top of the network. In the biomedical domain,
\newcite{Habibi2017} used this architecture for chemical and gene name recognition. 
\newcite{liu2017identification} and \newcite{Dernoncourt2017} adapted it for
state-of-the-art note deidentification. In terms of functioning, \newcite{Kukafka2006}
and \newcite{Skube2018} investigate the presence of functioning terminology in
clinical data, but do not evaluate it from an NER perspective.

%% file: sections/data.tex
\input{figures/synthetic_data}

\section{Data}
\label{sec:data}

\newcite{Thieu2017} presented a dataset of 250 de-identified EHR documents
collected from
Physical Therapy (PT) encounters at the Clinical Center of the National
Institutes of Health (NIH). These documents, obtained from the NIH Biomedical
Translational Research Informatics System (BTRIS; \citealt{BTRIS}), were
annotated for several aspects
of patient mobility, a subdomain of functioning-related activities defined by
the ICF; we therefore refer to this dataset as \dataset. We focus on two types
of contiguous text spans:
descriptions of mobility status, which we call Mobility entities, and
measurement scales related to mobility activity, which we refer to as
ScoreDefinition entities.

Two major differences stand out in \dataset\ as compared with standard NER
data. The entities, defined for this task as contiguous text spans completely
describing an aspect of mobility, tend to be quite long: while prior NER datasets
such as the i2b2/VA 2010 shared task data \cite{Uzuner2012} include fairly
short entities (2.1 tokens on average for i2b2), \Mob\ entities are an average of 10
tokens long, and \ScoreDef\ average 33.7 tokens. Also,
both \Mob\ and \ScoreDef\ entities tend to be entire clauses or sentences,
in contrast with the constituent noun phrases that are the meat of most NER.
Figure~\ref{fig:synthetic-data} shows example \Mob\ and
\ScoreDef\ entities in a short synthetic document. Despite these challenges,
\newcite{Thieu2017} show high ($>0.9$) inter-annotator agreement on the text
spans, supporting use of the data for training and evaluation.

These characteristics align well with past successful applications of
recurrent neural models to challenging NLP problems. For our evaluation
on this dataset, we randomly split \dataset\  at document level into
training, validation, and test sets, as described in Table~\ref{tbl:dataset}.

\input{tables/dataset}

\subsection{Text corpora}

In order to learn input word embeddings for NER, we
use a variety of both in-domain and out-of-domain corpora, defined in terms
of whether the corpus documents include descriptions of function.
For in-domain data, with explicit references to patient functioning, we use
a corpus of
154,967 EHR documents shared with us (under an NIH Clinical Center Office of
Human Subjects determination) from the NIH BTRIS system.\footnote{
    There is no overlap between these documents and the annotated data
    in \dataset\ (T.\ Thieu, personal communication).
}
A large
proportion of these documents comes from the Rehabilitation Medicine Department of the
NIH Clinical Center, including Physical Therapy (PT), Occupational Therapy
(OT), and other therapeutic records; the remaining documents
are sampled from other departments of the Clinical Center.

Since \dataset\ is focused on PT documents, we also use a subset of this corpus
consisting of 17,952 PT and OT documents. Despite this small size, the topical similarity of these documents makes
them a very targeted in-domain corpus. For clarity, we refer to the full
corpus as \BTRIS, and the smaller subset as \PTOT.

\subsubsection{Out-of-domain corpora}

As the \BTRIS\ corpus is considered a small training corpus for
learning word embeddings, we also use three larger out-of-domain corpora, which
represent different degrees of difference from the in-domain data.
Our largest data source is pretrained FastText embeddings
from Wikipedia 2017, web crawl data, and news documents.\footnote{
    {\footnotesize \texttt{fasttext.cc/docs/en/english-vectors}}
}

We also make use of two biomedical corpora for comparison with existing
work. PubMed abstracts have been an extremely useful source of embedding
training in biomedical NLP \cite{Chiu2016b}; we use the text of approximately
14.7 million abstracts taken from the 2016 PubMed baseline as a
high-resource biomedical corpus. In addition, we use two million
free-text documents released as part of the
MIMIC-III critical care database \cite{Johnson2016}. Though smaller than
PubMed, the MIMIC corpus is a large sample of clinical text, which is
often difficult to obtain and shows significant linguistic differences
with biomedical literature \cite{Friedman2002}. As MIMIC is clinical text,
it is the closest comparison corpus to the \BTRIS\ data; however, as MIMIC
focuses on ICU care, the information in it differs significantly from
in-domain \BTRIS\ documents.

%% file: figures/synthetic_data.tex
\begin{figure}[t]
    \begin{quote}
        \tt \small
        Evaluation:

        \SDColor{[Scoring:\ 1=totally dependent, 2=requires assistance,
        3=requires appliances, 4=totally independent]$_{\textrm{\ScoreDef}}$}.

        \MobColor{[Ambulation:\ 4]$_{\textrm{\Mob}}$}\\
        Observations:\\
        Pt is weight bearing:\ \MobColor{[she ambulates independently w/o
        use of assistive device]$_{\textrm{\Mob}}$}.\\
        Limited to very brief examination.
    \end{quote}
    \setlength{\abovecaptionskip}{-7pt}
    \setlength{\belowcaptionskip}{-8pt}
    \caption{Synthetic document with examples of \ScoreDef\ (in blue) and \Mob\ (in orange).}
    \label{fig:synthetic-data}
\end{figure}

%% file: tables/dataset.tex
\begin{table}[t]
    \centering
    \begin{tabular}{l|c|c|c}
        Entity&Train&Valid&Test\\
        \hline
        \Mob&1,533&467&947\\
        \ScoreDef&82&24&48\\
    \end{tabular}
    \caption{Named entity statistics for training, validation, and test splits
             of \dataset. Due to the rarity of \ScoreDef\ entities, we use a
             2:1 split of training to test data, and hold out 10\% of training
             data as validation.}
    \label{tbl:dataset}
\end{table}

%% file: sections/methods.tex
\section{Methods}

\begin{figure}[t]
	\centering
	\scalebox{0.43}{
		\includegraphics[]{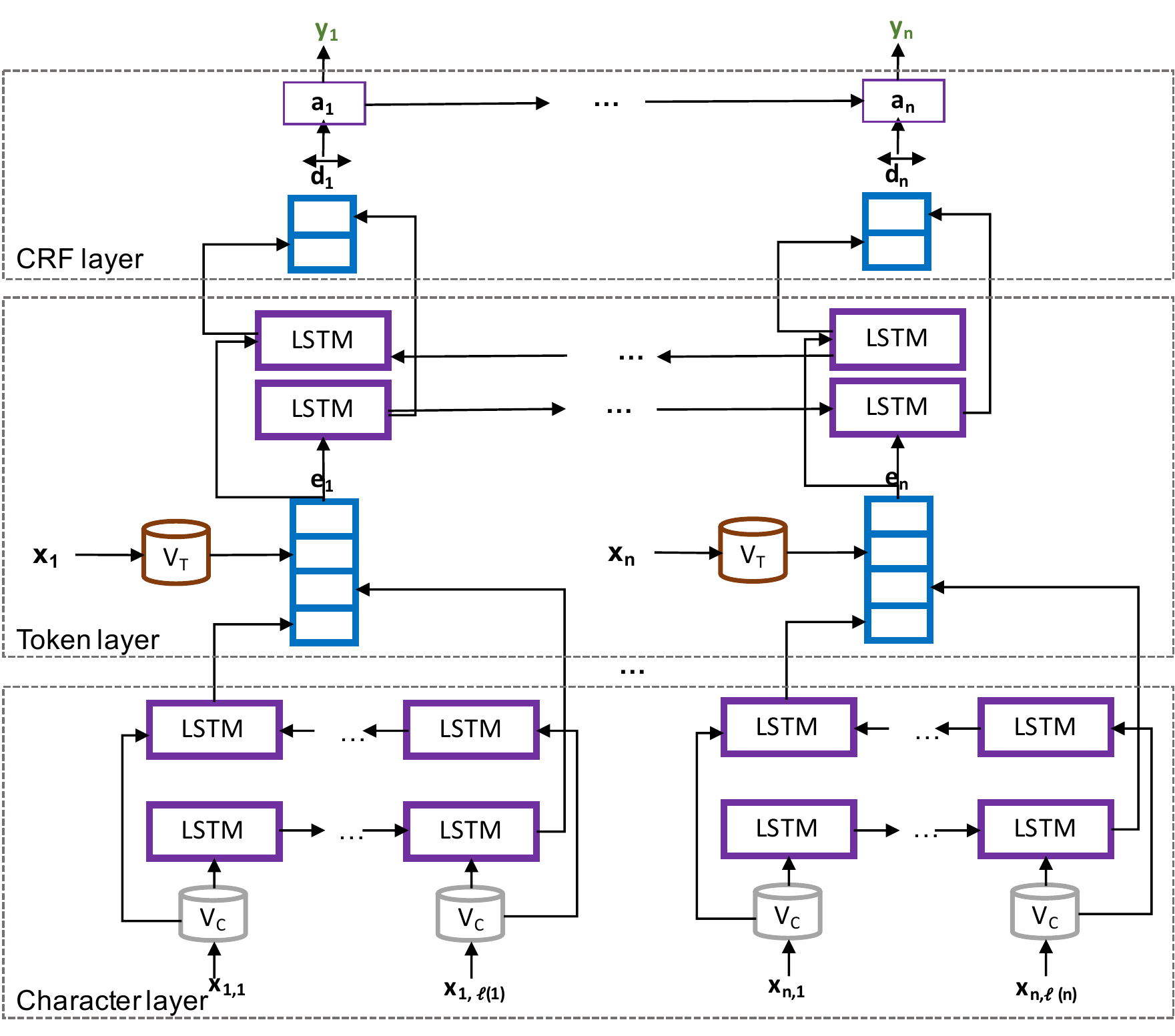}}
    \setlength{\abovecaptionskip}{-5pt}
	\setlength{\belowcaptionskip}{-12pt}
	\caption{Bi-LSTM-CRF network architecture}
	\label{fig:bi_lstm}
\end{figure}

We adopt the architecture of \newcite{Dernoncourt2017}, due to its successful
NER results on CoNLL and i2b2 datasets. The architecture, as depicted in
Figure~\ref{fig:bi_lstm}, is a stacked LSTM composed of:\ i) character Bi-LSTM
layer that generates character embeddings. We include this in our experimentations
due to its performance enhancement; ii) token Bi-LSTM layer
using both character and pre-trained word embeddings as input; iii) CRF layer
to enhance the performance by taking into account the surrounding
tags~\citep{lample2016neural}.
We use the following values for the network hyperparameters, as they yielded
the best performance on the validation set:
i) hidden state dimension of~25 for both character and token layers. In contrast to more common token layer sizes such as 100 or 200, we found the best validation set performance for our task with 25 dimensions; 
ii) learning rate =~0.005; iii) patience =~10;
iv) optimization with stochastic gradient descent (SGD) which showed superior performance to adaptive moment estimation (Adam) optimization technique~\cite{adam}.         

\subsection{Embedding training}

We use two popular toolkits for learning word embeddings: word2vec\footnote{
    We use word2vec modified to support pre-initialization, from
    {\tt github.com/drgriffis/word2vec-r}.
} \cite{Mikolov2013a} and FastText\footnote{
    {\footnotesize \texttt{github.com/facebookresearch/fastText}}
} \cite{Bojanowski2017}. We run both toolkits using skip-gram with negative sampling
to train 300-dimensional embeddings, and use default settings for all other
hyperparameters.\footnote{
    For \PTOT\ embeddings, due to the extremely small corpus size, we use an initial learning
    rate of 0.05, keep all words with minimum frequency 2, and train for 25 iterations.
}

\input{tables/unmapped_results}

\subsection{Domain adaptation methods}

We evaluate several different methods for adapting out-of-domain embeddings to
the \BTRIS\ corpus.

\textbf{Concatenation} In addition to the original embeddings, we
concatenate out-of-domain and \BTRIS/\PTOT\ embeddings as a baseline, allowing
the model to learn a task-specific combination of the two representations.

\textbf{Preinitialization} Recent work has found benefits from retraining learned embeddings on a target
corpus~\cite{yang2017simple}. We pre-initialize both word2vec and FastText toolkits
with embeddings learned on each of our three reference corpora, and retrain on the
\BTRIS\ corpus using an initial learning rate of 0.1. Additionally,
we use the regularization-based domain adaptation approach introduced
by~\newcite{yang2017simple} as another baseline, due to its successful results in
improving NER performance. Their method aims to help the model to differentiate between
general and domain specific terms, using a significance function $\phi$ of a word $w$.
$\phi$ is dependent on the definition of $w$'s frequency, where in our implementation it is the word frequency in the target corpora. 

\textbf{Linear transform} However,
these approaches suffer from the same limitations as training \BTRIS\ embeddings
directly: a restricted vocabulary and minimal training data, both due to the
size of the corpus. We therefore also investigate two methods for learning a
transformation from one set of embeddings into the same space as another,
based on a reference dictionary. Given an out-of-domain source
embedding set and a target \BTRIS\ embedding set, we use all words in common
between source and target as our training vocabulary.\footnote{
    We evaluated using subsets of 1k, 2k, or 10k shared words most frequent
    in \BTRIS,
    but the best
    downstream performance was achieved using all pivot points.
} We adapt this to the linear transformation method successfully applied to bilingual
embeddings by \newcite{Artetxe2016}, using this shared vocabulary as the training
dictionary.

\textbf{Non-linear transform} As all of our embeddings are in English, but from domains
that do not intuitively seem to have a linear relationship, we also extend the method
of Artetxe et al.\ to a non-linear transformation.
We randomly divide the shared vocabulary into ten folds, and train
a feed-forward neural network using nine-tenths of the data, minimizing mean squared error
(MSE) between the learned projection and the true embeddings. After each epoch, we calculate
MSE on the held-out set, and halt when this error stops decreasing. Finally, we average
the learned projections from each fold to yield the final transformation function.
Following \newcite{Artetxe2016}, we apply this function to all source embeddings, allowing
us to maintain the original vocabulary size.

Our model is a fully-connected feed-forward neural network, with the same hidden dimension
as our embeddings. We evaluate with both 1 and 5 hidden layers, and use either tanh
or rectified linear unit (ReLU) activation throughout. Model structure is denoted in the
result; for example, ``5-layer ReLU'' refers to nonlinear mapping using a 5-layer network with
ReLU activation. We train with Adam optimization \cite{adam} and a minibatch size of 5.\footnote{
    Source implementation available at\\
    {\tt github.com/drgriffis/NeuralVecmap}
}

%% file: tables/unmapped_results.tex
\begin{table*}[t]
    \centering
    {\small
    \begin{tabular}{l|c|c|ccc|ccc|ccc|ccc}
        \multirow{3}{*}{Corpus}&\multirow{3}{*}{Size}&\multirow{3}{*}{Toolkit}&\multicolumn{6}{c|}{\Mob}&\multicolumn{6}{c}{\ScoreDef}\\
        &&&\multicolumn{3}{c|}{Exact match}&\multicolumn{3}{c|}{Token match}&\multicolumn{3}{c|}{Exact match}&\multicolumn{3}{c}{Token match}\\
        &&&Pr&Rec&F1&Pr&Rec&F1&Pr&Rec&F1&Pr&Rec&F1\\
        \hline
        \multicolumn{3}{c|}{\it Random initialization}&
            67.7&61.8&64.6&84.0&75.9&79.7&
            86.5&93.4&90.0&97.7&98.9&98.3\\
        \hline
        WikiNews&16B&FT&
            67.0&64.0&65.4&83.0&80.0&81.5&
            83.3&93.4&88.2&96.8&99.3&98.0\\
        PubMed&2.6B&FT&
            68.7&\textbf{65.9}&67.2&82.0&84.5&83.2&
            \textbf{93.6}&91.7&92.6&\textbf{98.1}&97.8&97.9\\
        &&w2v&
            64.9&64.7&64.8&77.4&\textbf{87.7}&82.2&
            90.0&93.8&91.8&97.8&99.6&\textbf{98.7}\\
        MIMIC&497M&FT&
            37.7&10.6&16.5&78.9&21.7&34.0&
            86.0&90.0&87.8&97.9&97.7&97.8\\
        &&w2v&
            \textbf{71.9}&64.9&\textbf{68.2}&84.3&83.0&\textbf{83.6}&
            91.7&91.7&91.7&96.5&99.6&98.0\\
        \hline
        \BTRIS&74.6M&FT&
            66.8&63.8&65.3&80.6&83.4&82.0&
            90.2&\textbf{95.8}&92.9&95.9&99.0&97.4\\
        &&w2v&
            69.7&63.7&66.7&\textbf{86.0}&79.2&82.4&
            88.2&93.8&90.9&96.7&\textbf{99.9}&98.3\\
        \PTOT&4.2M&FT&
            68.8&62.5&65.5&84.5&80.2&82.3&
            92.0&\textbf{95.8}&\textbf{93.9}&97.1&97.7&97.4\\
        &&w2v&
            70.8&63.4&67.0&85.8&79.4&82.5&
            86.3&91.7&88.9&96.3&98.9&97.6\\
    \end{tabular}
    }
    \caption{Exact and token-level match results on \dataset, using randomly-initialized
             embeddings as a baseline and unmodified word2vec (w2v) and FastText (FT)
             embeddings from different corpora.
             \textit{Size} is the number of tokens in the training corpus.}
    \label{tbl:unmapped-results}
\end{table*}

%% file: sections/results.tex
\section{Results}
\label{sec:results}
We report exact match results, calculated using CoNLL 2003 named
entity recognition shared task evaluation scoring~\cite{CoNLL03},
which requires that all tokens of an entity are correctly recognized.
Additionally, given the long span of \Mob\ and \ScoreDef\ entities
(see Section~\ref{sec:data}), we evaluated partial match performance
using token-level results. For simplicity, we report only performance
on the test set; however, validation set numbers consistently follow
the same trends observed in test data. We denote embeddings trained
using FastText with the subscript $_{FT}$, and word2vec with $_{w2v}$.

\subsection{Embedding corpora}
\label{ssec:embedding-corpora}
Exact and token-level match results for both \Mob\ and \ScoreDef\ entities are given for
embeddings from each corpus in Table~\ref{tbl:unmapped-results}.
By and large, the in-domain \BTRIS\ and \PTOT\ embeddings yield
higher precision than out-of-domain embeddings, though this comes at the
expense of recall. word2vec embeddings consistently achieve 
better NER performance than FastText embeddings from the clinical corpora,
although this was reversed with PubMed, suggesting that further
research is needed on the strengths of different embedding methods in
biomedical data. The unusually poor performance of MIMIC$_{FT}$
embeddings persisted across multiple experiments with two embedding
samples, manifesting primarily in making very few predictions (less
than 30\% as many \Mob\ entities other embeddings yielded).

Most notably, despite a thousand-fold reduction in training corpus size,
we see that \PTOT\ embeddings match the performance of PubMed
embeddings on \Mob\ mentions and achieve the best overall performance on
\ScoreDef\ entities. Together with the overall superior performance
of \PTOT\ embeddings even to the larger \BTRIS\ corpus, our findings support
the value of using input embeddings
that are highly representative of the target domain.  Nonetheless, MIMIC
embeddings have both the best precision and overall performance on \Mob\ 
data, despite the domain mismatch of critical care versus
therapeutic encounters. This indicates that there is a limit to the benefits
of in-domain data that can be outweighed by sufficient data from a different
but related domain.

Token-level results follow the same trends as exact match, with
clinical embeddings achieving highest precision, while PubMed embeddings
yield better recall. As many entity-level errors are only off by a
few tokens, token-level scores are generally 15-20 absolute points higher
than their corresponding entity-level scores.  At the token
level, it is clear that \ScoreDef\ entities are effectively solved in
this dataset, with all F1 scores are above 97.4\%. This is primarily
due to the regularity of \ScoreDef\ strings: they typically consist of a
sequence of single numbers followed by explanatory strings, as shown in
Figure~\ref{fig:synthetic-data}.

\setcounter{table}{3}
\input{tables/mapped_consistency}
\setcounter{table}{2}
\input{tables/mapped_results}
\setcounter{table}{4}

\subsection{Mapping methods}

Table~\ref{tbl:mapped-results} takes a single representative source/target
pair and compares the different results obtained on recognizing \Mob\ entities
when the NER model is initialized with embeddings learned using different domain
adaptation methods. In this case, as with several other source/target pairs we
evaluated, the concatenated embeddings give the best overall performance,
stemming largely from an increase in recall over the baselines. However, we
see that the nonlinear mapping methods tend to yield high precision: all settings
improve over WikiNews embeddings alone, and the 1-layer tanh mapping beats
the \BTRIS\ embeddings as well. Reflecting the earlier observed trends of
in-domain data, this is offset by a drop in recall, often of several absolute
percentage points.

These differences are fleshed out further in Table~\ref{tbl:mapped-consistency},
comparing four domain adaptation methods across several source/target pairs.
Concatenation typically achieves the best overall performance among the adaptation
methods, but nonlinear mappings yield highest precision in 6 of the 8 settings
shown.
Concatenation is also more sensitive to noise in the
source embeddings, as shown with MIMIC$_{FT}$ results, and preinitialization
varies widely in its performance. By contrast, linear and nonlinear mapping
methods are less affected by the choice of source embeddings,
yielding more consistent results than preinitialization or concatenation for
a given target corpus. Nonlinear mappings exhibit this stability most clearly,
producing very similar results across all settings. The regularization-based
domain adaptation method of
\newcite{yang2017simple} consistently yielded similar results to preinitialization: for
example, an F1 score of 65\% when PubMed$_{w2v}$ embeddings are adapted to
\BTRIS, as compared to 65.4\% using pre-initialization with word2vec. We 
therefore omit these results for brevity.

Comparing both Tables~\ref{tbl:mapped-results}~and~\ref{tbl:mapped-consistency}
to the performance of unmodified embeddings shown in
Table~\ref{tbl:unmapped-results}, we see a surprising lack of overall
performance improvement or degradation. While the different adaptation methods
exhibit consistent differences between one another, only 12 of the 32 F1 scores
in Table~\ref{tbl:mapped-consistency} represent improvements over the relevant unmapped
baselines. Many adaptation results achieve notable improvement in precision or
recall individually, suggesting that different methods may be more useful for
downstream applications where one metric is emphasized over the other. However,
several of our results indicate failure to adapt, illustrating the difficulty of
effectively adapting embeddings for this task.

\input{tables/best_results}

\subsection{Source/target pairs}

Table~\ref{tbl:best-results} highlights the source/target pairs that achieved
the best exact match precision, recall, and F1 out of all the embeddings we evaluated,
both unmapped and mapped. Though each source/target pair produced varying
downstream results among the domain adaptation methods, a couple of broad
trends emerged from our analysis. The largest performance gains over unmapped
baselines were found when adapting high-resource WikiNews and PubMed
embeddings to in-domain representations; however, these pairings also had
the highest variability in results. The most consistent gains in precision
came from using MIMIC embeddings as source, and these were mostly achieved
through the nonlinear mapping approach.

There was no clear trend in the
domain-adapted results as to whether word2vec or FastText
embeddings led to the best downstream performance: it varied between pairs
and adaptation methods. word2vec embeddings were generally more consistent,
but as seen in
Tables~\ref{tbl:mapped-consistency} and \ref{tbl:best-results}, FastText
embeddings often achieved the highest performance.

\subsection{Error analysis}

Several interesting trends emerge in the NER errors produced
in our experiments. Most generally,
punctuation is often falsely considered to bound an entity. For example,
the following string is part of a continuous \Mob\ entity:\footnote{
    Several examples in this section have been edited for
    deidentification purposes and brevity.
}
\begin{quoting}
    \tt \small supine in bed with elevated leg, and was left sitting in bed
\end{quoting}
However, most trained models separated this at the comma into two \Mob\ 
entities. Unsurprisingly, given the length of \Mob\ 
entities, we find many cases where most of the correct entity is tagged
by the model, but the first or last few words are left off, as in
\begin{quoting}
    \tt \small [he exhibits compensatory gait patterns]$_{Pred}$ as a
    result]$_{Gold}$
\end{quoting}
This behavior is illustrated in the large performance difference between
entity-level and token-level evaluation discussed in
Section~\ref{ssec:embedding-corpora}.

We also see that descriptions of physical activity without
specific evaluative terminology are often missed by the model. For
example, {\small \texttt{working out in the yard}} is a \Mob\ entity
ignored by the vast majority of our experiments, as is
{\small \texttt{negotiate six steps to enter the apartment}}.

\subsubsection{Corpus effects}

Within correctly predicted entities, we see some indications of source
corpus effect in the results. Considering just the original, non-adapted
embeddings as presented in Table~\ref{tbl:unmapped-results}, we note two
main differences between models trained on out-of-domain vs in-domain
embeddings. In-domain embeddings lead to much more conservative
models: for example, \PTOT$_{w2v}$ only predicts 850 \Mob\ entities in
test data, and \BTRIS$_{w2v}$ predicts 863; this is in contrast to 922
predictions from MIMIC$_{w2v}$ and 940 from PubMed$_{w2v}$. This carries
through to mapped embeddings as well: adding \PTOT\ embeddings into the
mix decreases the number of predictions across the board.

Several predictions exhibit some degree of domain sensitivity,
as well. For example, ``fatigue'' is present at the end of several
\Mob\ mentions, and both PubMed and MIMIC embeddings typically end
these mentions early. PubMed embeddings also append more
typical symptomatic language onto otherwise correct \Mob\ entities,
such as
{\small \texttt{no areas of pressure noted on skin}} and
{\small \texttt{numbness and tingling of arms}}. MIMIC and the
heterogeneous in-domain \BTRIS\ corpus append
similar language, including {\small \texttt{and chronic pain}}.
WikiNews embeddings, by contrast, appear oversensitive to key words
in many \Mob\ mentions, tagging false positives such as
{\small \texttt{my wife}} (spouses are often referred to
as a source of physical support) and {\small
\texttt{stairs are within range}}.

\input{figures/nn_changes}
\input{tables/nn_examples}

\subsubsection{Changes from domain adaptation}

Domain-adapted embeddings fix some corpus-based issues,
but re-introduce others. Out-of-domain corpora tend to chain together
\Mob\ entities separated by only one or two words, as in
\begin{quoting}
    \tt \small \MobColor{[He ambulates w/o ad]$_{\Mob}$}, no walker
    observed, \MobColor{[antalgic gait pattern]$_{\Mob}$}
\end{quoting}
While source PubMed and WikiNews embeddings often collapse these to
a single mention, adapting them to the target domain fixes many
such cases. However, some of the original corpus noise remains:
\PTOT$_{w2v}$ correctly ignored {\small \texttt{and
chronic pain}} after a \Mob\ mention, but MIMIC$_{w2v}$ mapped to
\PTOT$_{w2v}$ re-introduces
this error.

The most consistent improvement obtained from domain adaptation was
on \Mob\ entities that are short noun phrases, e.g. {\small
\texttt{gait instability}}, and {\small \texttt{unsteady gait}}. 
Non-adapted embeddings typically miss such phrases, but mapped
embeddings correctly find many of them, including some that
in-domain embeddings miss.

\subsubsection{Adaptation method effects}

The most striking difference we observe when comparing
different domain adaptation methods is that preinitialization universally
leads to longer \Mob\ entity predictions, by both mean and variance
of entity length. Though preinitialized embeddings still
perform well overall, many predictions include several extra tokens
before or after the true entity, as in the following example:
\begin{quoting}
    \tt \small (now that her leg is healed \MobColor{[she is independent with
    wheelchair transfer]$_{Gold}$} and using her shower bench)$_{Pred}$
\end{quoting}
Preinitialized embeddings also have a strong tendency to collapse sequential
\Mob\ entities. Both of these trends are
reflected in the lower token-level precision numbers in
Table~\ref{tbl:mapped-results}.

Comparing nonlinear mapping methods, we find that a 1-layer mapping with
tanh activation consistently leads to fewer predicted \Mob\ entities than
with ReLU (for example, 814 vs 859 with WikiNews$_{FT}$ mapped
to \BTRIS$_{w2v}$, 917 vs 968 with MIMIC$_{w2v}$ mapped to \PTOT$_{w2v}$).
However, this difference disappears when a 5-layer mapping is used. Despite
their consistent performance, nonlinear transformations seem to
re-introduce a number of errors related to more general
mobility terminology. For example, {\small \texttt{he is very active and runs
15 miles per week}} is correctly recognized by concatenated WikiNews$_{FT}$
and \BTRIS$_{w2v}$, but missed by several of their nonlinear mappings.

%% file: tables/mapped_consistency.tex
\begin{table*}[t]
    \centering
    {\small
        \begin{tabular}{cl|ccc|ccc|ccc|ccc}
            \multirow{2}{*}{Target}&\multirow{2}{*}{Source}&\multicolumn{3}{c|}{Concat}&\multicolumn{3}{c|}{Preinit}&\multicolumn{3}{c|}{Linear}&\multicolumn{3}{c}{5-layer tanh}\\
            &&Pr&Rec&F1&Pr&Rec&F1&Pr&Rec&F1&Pr&Rec&F1\\
            \hline
            \multirow{5}{*}{\BTRIS$_{FT}$}&WikiNews$_{FT}$&
                \textbf{72.2}&\textbf{65.3}&\textbf{68.6}&
                55.0&59.2&57.0&
                65.1&61.9&63.5&
                69.3&64.2&66.7\\
            &PubMed$_{FT}$&
                \textbf{69.5}&65.8&\textbf{67.6}&
                64.2&\textbf{66.5}&65.4&
                65.6 &60&62.7&
                66.1&64.5&65.3\\
            &PubMed$_{w2v}$&
                65.3&65.3&\textbf{65.3}&
                64.8&\textbf{65.4}&65.1&
                70.3&65.8&68&
                \textbf{66.3}&62.6&64.4\\
            &MIMIC$_{FT}$&
                35.0&10.4&16.0&
                37.8&15.5&22.0&
                63.7&\textbf{62.9}&63.3&
                \textbf{70.3}&61.3&\textbf{65.5}\\
            &MIMIC$_{w2v}$&
                67.4&\textbf{67.6}&\textbf{67.5}&
                68.5&64.6&66.5&
                66.8&60.3&63.4&
                \textbf{69.2}&64.3&66.7\\
            \hline
            \multirow{3}{*}{\PTOT$_{FT}$}&WikiNews$_{FT}$&
                67.5&\textbf{63.9}&65.6&
                54.5&57.9&56.1&
                68.9&63.8&66.2&
                \textbf{68.5}&63.4&\textbf{65.8}\\
            &PubMed$_{FT}$&
                62.8&\textbf{65.1}&\textbf{63.9}&
                61.3&50.2&55.2&
                62.6&62.6&62.6&
                \textbf{68.3}&60.1&\textbf{63.9}\\
            &MIMIC$_{w2v}$&
                64.1&\textbf{66.1}&\textbf{65.1}&
                59.9&61.8&60.8&
                57.9&54.1&55.9&
                \textbf{67.3}&63.2&\textbf{65.1}\\
        \end{tabular}
    }
    \caption{Exact match precision and recall for \Mob\ entities with word
             embeddings mapped
             from each source to \BTRIS$_{FT}$ embeddings, using four
             selected domain
             adaptation methods. The
             best-performing embeddings from each source corpus were also
             mapped to \PTOT$_{FT}$ embeddings. The best precision,
             recall, and F1 achieved with each source/target pair is marked in bold.}
    \label{tbl:mapped-consistency}
\end{table*}

%% file: tables/mapped_results.tex
\begin{table}[t]
    \centering
    {\small
    \setlength{\tabcolsep}{5pt}
    \begin{tabular}{l|ccc|ccc}
        \multirow{2}{*}{Method}&\multicolumn{3}{c|}{Exact match}&\multicolumn{3}{c}{Token match}\\
        &Pr&Rec&F1&Pr&Rec&F1\\
        \hline
        WikiNews$_{FT}$   &67.0&64.0&65.4&83.0&80.0&81.5\\
        \BTRIS$_{w2v}$    &70.0&63.7&66.6&\textbf{86.0}&79.2&81.5\\
        \hline
        Concatenated  &68.6&\textbf{66.7}&\textbf{67.6}&84.3&81.8&\textbf{83.0}\\
        Preinitialized&66.8&64.5&65.6&78.4&\textbf{86.4}&82.2\\
        \hline
        Linear        &\textbf{72.5}&58.9&65&79.1&83&81\\
        1-layer ReLU  &69.2&63.2&66.0&83.4&76.9&80.0\\
        1-layer tanh  &70.6&61.0&65.5&84.9&75.7&80.1\\
        5-layer ReLU  &67.3&61.9&64.5&83.5&76.6&79.9\\
        5-layer tanh  &67.9&62.1&64.9&82.1&77.0&79.4\\
    \end{tabular}
    }
    \setlength{\belowcaptionskip}{-12pt}
    \caption{Comparison of mapping methods, using WikiNews$_{FT}$
             as source and \BTRIS$_{w2v}$ as target.
             Results are given for exact
             entity-level match and token-level match for test set
             \Mob\ entities.}
    \label{tbl:mapped-results}
\end{table}

%% file: tables/best_results.tex
\begin{table}[t]
    \centering
    {\small
    \setlength{\tabcolsep}{4pt}
    \begin{tabular}{l|l|l|ccc}
        Source&Target&Method&Pr&Rec&F1\\
        \hline
        WikiNews$_{FT}$&\PTOT$_{w2v}$&Preinit&72.1&66.1&\textbf{69.0}\\
        WikiNews$_{FT}$&\BTRIS$_{w2v}$&Linear&\textbf{72.5}&58.9&65\\
        MIMIC$_{w2v}$&\BTRIS$_{FT}$&Concat&67.4&\textbf{67.6}&67.5\\
    \end{tabular}
    }
    \caption{Best precision, recall, and F1 (exact) for test set \Mob\ mentions, with
             the source/target pair and domain adaptation method used.}
    \label{tbl:best-results}
\end{table}

%% file: figures/nn_changes.tex
\begin{figure*}[t]
    \centering
    \includegraphics[width=\textwidth]{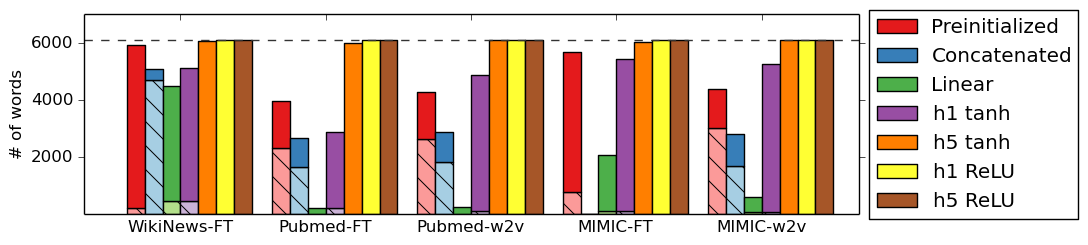}
    \setlength{\abovecaptionskip}{-12pt}
    \caption{Number of words in shared vocabulary with different nearest
             neighbors in source and domain-adapted embeddings, using
             \BTRIS$_{FT}$ as target. Light hatched bars indicate the number
             of words whose new nearest neighbor matches
             \BTRIS$_{FT}$. The dashed line indicates
             shared vocabulary size.}
    \label{fig:nn-changes}
\end{figure*}

%% file: tables/nn_examples.tex
\begin{table*}[t]
    \centering
    {\small
    \begin{tabular}{|c|cc|ccccc|}
        \hline
        Source set&Source     &Target      &Preinit   &Concat     &Linear&h1 tanh   &h5 tanh\\
        \hline
        \multirow{3}{*}{PubMed$_{FT}$}
            &ambulating &ambulating  &ambulating&ambulating &ambulating     &ambulating&worsening   \\
            &ambulate   &ambulate    &ambulate  &ambulate   &ambulate       &ambulate  &wearing     \\
            &crutches   &ambulatory  &walker    &crutches   &crutches       &crutch    &complaints  \\
        \hline
        \multirow{3}{*}{WikiNews$_{FT}$}
            &ambulating &ambulating  &pos       &ambulating &cardiopulmonary&robotic   &respiratory \\
            &ambulate   &ambulate    &76        &ambulate   &neurosurgical  &overhead  &sclerotic   \\
            &extubation &ambulatory  &acuity    &ambulatory &resuscitation  &ambulating&acupuncture \\
        \hline
    \end{tabular}
    }
    \caption{Top 3 nearest neighbors of \textit{ambulation} in embeddings mapped
             to \BTRIS$_{FT}$ using different adaptation methods.  Source and
             Target are neighbors in the original source and \BTRIS$_{FT}$
             embeddings.}
    \label{tbl:nn-examples}
\end{table*}

%% file: sections/qualitative.tex
\section{Embedding analysis}

To further evaluate the effects of different domain adaptation methods, we
analyzed the nearest neighbors by cosine similarity of each word before and
after domain
adaptation. We only considered the words present both in the dataset and in
each of our original sets of embeddings, yielding a vocabulary of 6,201 words.
We then took this vocabulary and calculated nearest neighbors within it,
using each set of out-of-domain original embeddings and each of its
domain-adapted transformations.

Figure~\ref{fig:nn-changes} shows the number
of words whose nearest neighbors changed after adaptation, using \BTRIS$_{FT}$
as the target; all other targets display similar results. We see that in
general, the neighborhood structure of target embeddings is well-preserved
with concatenation, sometimes preserved with preinitialization, and completely
disposed of with the nonlinear transformation. Interestingly, this
reorganization of words to something different from both source and target
does not lead to the performance degradation we might expect, as shown in
Section~\ref{sec:results}.

We also qualitatively examined nearest neighbors before and after
adaptation. Table~\ref{tbl:nn-examples} shows nearest neighbors of
\textit{ambulation}, a common \Mob\ word, for two
representative source/target pairs. Preinitialization generally
reflects the neighborhood structure of the target embeddings, but
can be noisy: in WikiNews$_{FT}$/\BTRIS$_{FT}$, other words such as
\textit{therapy} and \textit{fatigue} share \textit{ambulation}'s
less-than-intuitive neighbors.

Reflecting the changes seen in Figure~\ref{fig:nn-changes}, the linear
transformation preserves source neighbors in the biomedical PubMed corpus,
but yields a neighborhood structure different from source or target
with highly out-of-domain WikiNews embeddings. Nonlinear
transformations sometimes yield sensible nearest neighbors, as in the
single-layer tanh mapping of PubMed$_{FT}$ to \BTRIS$_{FT}$. More often,
however, the learned projection significantly shuffles neighborhood structure,
and observed neighbors may bear only a distant similarity to the query term.
In several cases, large swathes of the vocabulary are mapped to a single tight
region of the space, yielding the same nearest neighbors for
many disparate words. This occurs more often when using a ReLU activation,
but we also observe it occasionally with tanh activation.

%% file: sections/conclusions.tex
\section{Conclusions}

We have conducted an experimental analysis of recognizing descriptions of
patient mobility with a recurrent neural network, and of the effects of various
domain adaptation methods on recognition performance. We find that a state-of-the-art
recurrent neural model is capable of capturing long, complex descriptions of
mobility, and of recognizing mobility measurement scales nearly perfectly. Our
experiments show that domain adaptation methods often improve recognition
performance over both in- and out-of-domain baselines, though such improvements
are difficult to achieve consistently. Simpler methods such
as preinitialization and concatenation achieve better performance gains, but
are also susceptible to noise in source embeddings; more complex methods
yield more consistent performance, but with practical downsides such as
decreased recall and a non-intuitive projection of the embedding space. Most
strikingly, we see that embeddings trained on a very small corpus of highly
relevant documents nearly match the performance of embeddings trained on
extremely large out-of-domain corpora, adding to the recent findings of
\newcite{Diaz2016}.

To our knowledge, this is the first investigation into automatically recognizing descriptions
of patient functioning.
Viewing this problem through an NER lens provides a robust framework for
model design and evaluation, but is accompanied by challenges such as effectively
evaluating recognition of long text spans and dealing with complex syntactic
structure and punctuation within relevant mentions. It is our hope that
these initial findings, along with further research refining the appropriate
framework for representing and approaching the recognition problem,
will spur further research into this complex and important domain.

%% file: sections/acknowledgments.tex
\section*{Acknowledgments}

The authors would like to thank Elizabeth Rasch, Thanh Thieu, and Eric
Fosler-Lussier for helpful discussions, the NIH Biomedical Translational
Research Information System (BTRIS) for their support, and our anonymous
reviewers for their invaluable feedback. This research was
supported in part by the Intramural Research Program of the National
Institutes of Health, Clinical Research Center and through an Inter-Agency
Agreement with the US Social Security Administration.